\documentclass[conference]{IEEEtran}
\IEEEoverridecommandlockouts
\usepackage{cite}
\usepackage{amsmath,amssymb,amsfonts}
\usepackage{algorithmic}
\usepackage{algorithm}
\usepackage{graphicx}
\usepackage{textcomp}
\usepackage{xcolor}
\usepackage{footnote}
\def\BibTeX{{\rm B\kern-.05em{\sc i\kern-.025em b}\kern-.08em
    T\kern-.1667em\lower.7ex\hbox{E}\kern-.125emX}}

\DeclareMathOperator*{\argmax}{arg\,max}
\DeclareMathOperator*{\argmin}{arg\,min}

\usepackage{adjustbox}

\def\BibTeX{{\rm B\kern-.05em{\sc i\kern-.025em b}\kern-.08em
    T\kern-.1667em\lower.7ex\hbox{E}\kern-.125emX}}

\begin{document}
	
	\title{Towards Modern Card Games with Large-Scale Action Spaces Through Action Representation\\ % through Deep Reinforcement Learning
		\thanks{© 2022 IEEE.  Personal use of this material is permitted.  Permission from IEEE must be obtained for all other uses, in any current or future media, including reprinting/republishing this material for advertising or promotional purposes, creating new collective works, for resale or redistribution to servers or lists, or reuse of any copyrighted component of this work in other works.}
	}

\author{\IEEEauthorblockN{Zhiyuan Yao\IEEEauthorrefmark{1},
Tianyu Shi\IEEEauthorrefmark{2}, Site Li\IEEEauthorrefmark{3}, Yiting Xie\IEEEauthorrefmark{3}, Yuanyuan Qin\IEEEauthorrefmark{3}, Xiongjie Xie\IEEEauthorrefmark{3}, Huan Lu\IEEEauthorrefmark{3} and
Yan Zhang\IEEEauthorrefmark{3}}
\IEEEauthorblockA{\IEEEauthorrefmark{1}School of Business, Stevens Institute of Technology, Hoboken NJ USA\\
\IEEEauthorrefmark{2}Toronto Intelligent Transportation Systems Centre, University of Toronto, Ontario Canada\\ \IEEEauthorrefmark{3}Deterrence, rct AI, Burbank CA USA\\
Email: \IEEEauthorrefmark{1}zyao9@stevens.edu,
\IEEEauthorrefmark{2}ty.shi@mail.utoronto.ca,
\IEEEauthorrefmark{3}lisite, xieyiting, qinyuanyuan, eric, hiker, yan@rct.ai,
}}

% 	\author{\IEEEauthorblockN{Anonymous Authors}}
% 	\author{\IEEEauthorblockN{Zhiyuan Yao}
% 		\IEEEauthorblockA{\textit{} \\
% 			\textit{RCT}\\
% 			Beijing, China \\
% 			@rct.ai}
% 		\and
% 		\IEEEauthorblockN{ABC*}
% 		\IEEEauthorblockA{\textit{RCT} \\
% 			\textit{RCT}\\
% 			Beijing, China \\
% 			@rct.ai}

% 	}
	
	\maketitle
	
	\begin{abstract}
	
% 	what is axie infinity, why we use this game? the challenge of high dimension space, why we use action encoding, 
	
% 	the performance comparison to other baselines.
	Axie infinity is a complicated card game with a huge-scale action space. This makes it difficult to solve this challenge using generic \textit{Reinforcement Learning} (RL) algorithms. We propose a hybrid RL framework to learn action representations and game strategies. To avoid evaluating every action in the large feasible action set, our method evaluates actions in a fixed-size set which is determined using action representations. We compare the performance of our method with the other two baseline methods in terms of their sample efficiency and the winning rates of the trained models. We empirically show that our method achieves an overall best winning rate and the best sample efficiency among the three methods.

	\end{abstract}
	
	\begin{IEEEkeywords}
		Game AI, Reinforcement Learning, Large-Scale Action Space, Action Representation, Axie Infinity
	\end{IEEEkeywords}

\section{Introduction}
% significance of our work?
% reinforcement learning has many successful applications in games. Card game as one typical class of the imperfect information game,  but large discrete action space in card game 
% Games is an ideal type of tasks for \textit{Reinforcement learning} (RL) studies, because the system dynamics of games are usually clearly defined, and the data of game can be easily populated to train RL models.
Games have facilitated the rapid development of RL algorithms in recent years.
% The success of Alpha Go \cite{silver2017mastering} and Alpha Zero  \cite{silver2018general} shows the RL-driven agents can achieve super-human performance on complex tasks like Go. 
% For instance, the well-known RL algorithm, Deep Q-leaning (DQN) \cite{mnih2015human}, produces models which performs human-level control on Atari video games. 
% The AI players of StarCraft II \cite{vinyals2017starcraft} demonstrate RL can handle complex multi-agent system. 
Card games, as a classical type of games, also pose many challenges to RL algorithms. 
% The direct applications of the generic RL algorithms have poor performance in many card games.
% One reason is that most of the cards games have large discrete action space, . 
The direct applications of generic algorithms \cite{mnih2015human, lillicrap2015continuous, mnih2016asynchronous, schulman2017proximal} in card games are problematic in many aspects because of the large-scale discrete action space \cite{zha2021douzero}. 
% for discrete action space problems like DQN\cite{mnih2015human} suffer from insufficient exploration in the large action space, which is formed by all possible permutations of the cards. 
% The continuous-action RL algorithms  need proper action abstraction to map the discrete action space to a continuous one. 
Prior works have proposed RL methods to approach a number of traditional card games, like Texas Hold'em \cite{bowling2015heads, brown2018superhuman, brown2019superhuman}, Mahjong \cite{li2020suphx}, DouDizhu \cite{zha2021douzero, guan2022perfectdou}, etc. However, the issues brought by the large action space still remain, especially for modern card games such as \textit{Axie Infinity} \footnote{ A detailed description of Axie Infinity can be found in  \url{https://whitepaper.axieinfinity.com/}} which has a huge discrete action space. 
% In card games like Poker, the agent has to deal with the imperfect-information, such as the unknown
% cards of the opponent~\cite{heinrich2016deep,he2016opponent}. In StarCraft II ~\cite{vinyals2017starcraft} , the
% agent has to compete with other agents who take simultaneous
% % actions from a large action space. 
% % Therefore, it is important to design an algorithm which can incorporate large discrete action space in such applications. 

Axie Infinity is an one-versus-one online card game which has millions of players globally.  \textit{Axies} are virtual pets that have different attributes such as species, health, speed, etc. Each axie has its own card deck consisting of 2 copies of 4 cards. The player needs to form a team of 3 axies at the beginning and play their cards (24 cards in total) to beat the opponent player's team.

This game is different from the traditional card game in the following aspects:
% Axie Infinity is a pay-to-earn game, in which the player buy NFTs and play, then produce more NFTs and make profit by selling it~\cite{dowling2022non}. Especially, it is  a modern card game which has millions (?) of global players. This is a imperfect two-player game. This game requires the player to form a team of 3 axies (?) and beat other players' team by playing cards. Each team can have very different strategies because the available cards are associated with the team lineup. This game brings many major challenges to RL studies because 
\begin{enumerate}
    \item For a fixed team, the player needs to choose one sequence of cards from 23 million different card sequences.
    % \item the feasible action sets are associated with the team status, and can be significantly different in different rounds. 
    \item The effect of a card is usually influenced by the its position in the card sequence and the status of the axies such as the health and the shield. 
    % \item Many factors have impact on the effect of each card, such as the other cards which are played with this card, the status of the opponent axies, etc.
    \item There are thousands of teams for players to choose. The optimal strategies for different teams are various.
    % It is difficult to gather the prior knowledge, as even an experienced player needs to learn from scratch when playing new teams. 
    % \item The available cards are random. The available cards are randomly chosen from the card deck. 
\end{enumerate}
These difficulties are all connected to the large-scale action space, and they are shared by a lot of other modern card games such as Hearthstone\footnote{\url{https://playhearthstone.com}}. 

Some existing works have investigated the large action space issue in card games. 
% These difficulties lead to the issue of knowledge generalization on actions -- the model should recognize the connections between actions. Also,
% The traditional ways to deal with discrete actions like one-hot encoding method are not suitable in this case because of the huge size. The action encoding should imply the connections between actions.
% Therefore, due to the complexity of the action space, it is unrealistic to let the algorithm recognize millions of action as independent discrete actions. 
% Briefly describe the task and challenges. 
% existing work on large action space
% One natural way to tackle these difficulties is to learn action representations. 
% guan2022perfectdou, zhao2022douzero+
Zha et al. \cite{zha2021douzero} propose an action encoding scheme for DouDizhu. However, this encoding scheme cannot properly encode the action in our problem as the complexity of the action space in our problem is much higher than that in DouDizhu (27472 possible moves). 
Dulac-Arnold et al. \cite{dulac2015deep} propose to choose actions in a small subset of the action space to speed up the action search process. This set is chosen based on a proper action encoding method which usually relies on prior knowledge. However, the prior human knowledge for our problem is hard to obtain due to the diversity of the teams and the strategies. Chandak et al. \cite{chandak2019learning} propose an algorithm to learn action representations from the consequences of corresponding actions. This method can avoid using prior human knowledge, but the policy-based method produces optimal actions which are not feasible in the discrete action set. 
% Action representation learning is a type of methods which learns a function to map the raw discrete action space to a latent action space. 
% action representation
% Existing works have investigated the issue brought by large discrete action space. In  
% and it usually only consists of a small portion of the original action space. 
%other action space shape techniques
 We also mention several general techniques in ~\cite{kanervisto2020action}  to reduce the size of the action space to improve the performance.  However, they are not applicable in our case as it still requires prior human knowledge of this game.
% Van et al. \cite{van2020q} improved the performance of Q-learning in large action space using a novel maximization operator. 

% Due to all the difficulties in Axie Infinity, the RL method for this problem should 1) quickly train models on different teams with minimal prior knowledge, 2) choose the optimal action only in a small subset of the large-scale feasible action set.

% two actions which have similar effects should have similar latent representations. RL algorithms can also benefit from the action representation learning due to efficient explorations in the continuous latent space.
% DouZero
% Our work is also inspired by those prior works on another card game -- DouDizhu. DouDizhu is a popular card game which is similar to Axie Infinity in many perspectives. A number of studies investigate the game DouDizhu from a RL perspective and propose some RL algorithms to train models for playing DouDizhu \cite{zha2021douzero, guan2022perfectdou, zhao2022douzero+}. Though numerous similarities are shared by two games, these works can neither be directly applied to Axie Infinity task. The reason is that Axie Infinity has a much larger action space, and the card deck for each team is also unique. It is therefore difficult to properly encode actions, and to generalize the knowledge on different teams.
In this paper, we consider a hybrid RL method to deal with the large discrete action space. This method chooses the optimal action in a small subset of the large-scale feasible action set. It can quickly train models on different teams with minimal prior knowledge. We test our method with other baseline methods using Axie Infinity task. We have the following main contributions:
\begin{enumerate}
    \item A Markov Decision Process (MDP) formulation for Axie Infinity problem including a novel action encoding scheme. 
    \item An efficient RL method to solve card game problems with large-scale discrete action space. 
    \item A supervised learning method to learn action representations. 
    % \item In the learning framework, we have also considered the team information to improve the learning efficiency 
    % \item An algorithm to pre-train the action embedding layer.
    \item Empirical results demonstrating the superiority than other baseline methods
\end{enumerate}

\section{Problem Formulation}\label{sec:formulation}
% There are thousands of teams in Axie Infinity. It is rather difficult to learn a universal model which understands how to play many teams since each team may have its unique strategy. Instead, learning to play a fixed team is easier. The trained agent should understand how to use a fixed team to maximize the winning chance when fighting against various different teams as the opponents. 

We assume our agent uses a fixed team to play against a rule-based player who randomly uses multiple popular teams. As the opponent is fixed, we formulate this problem as a single-agent Markov Decision Process (MDP). This MDP has a finite time horizon, each time step is one game round. The process is terminated when one player is defeated. The one-step transition probability measure is denoted as $P_t$. Every state in the state space $\mathcal{S}$ consists of all the available information to our player. This includes all six axies' status, energy, available cards, card history, etc.

\begin{figure}[h]
    \centering
    \includegraphics[width=8cm]{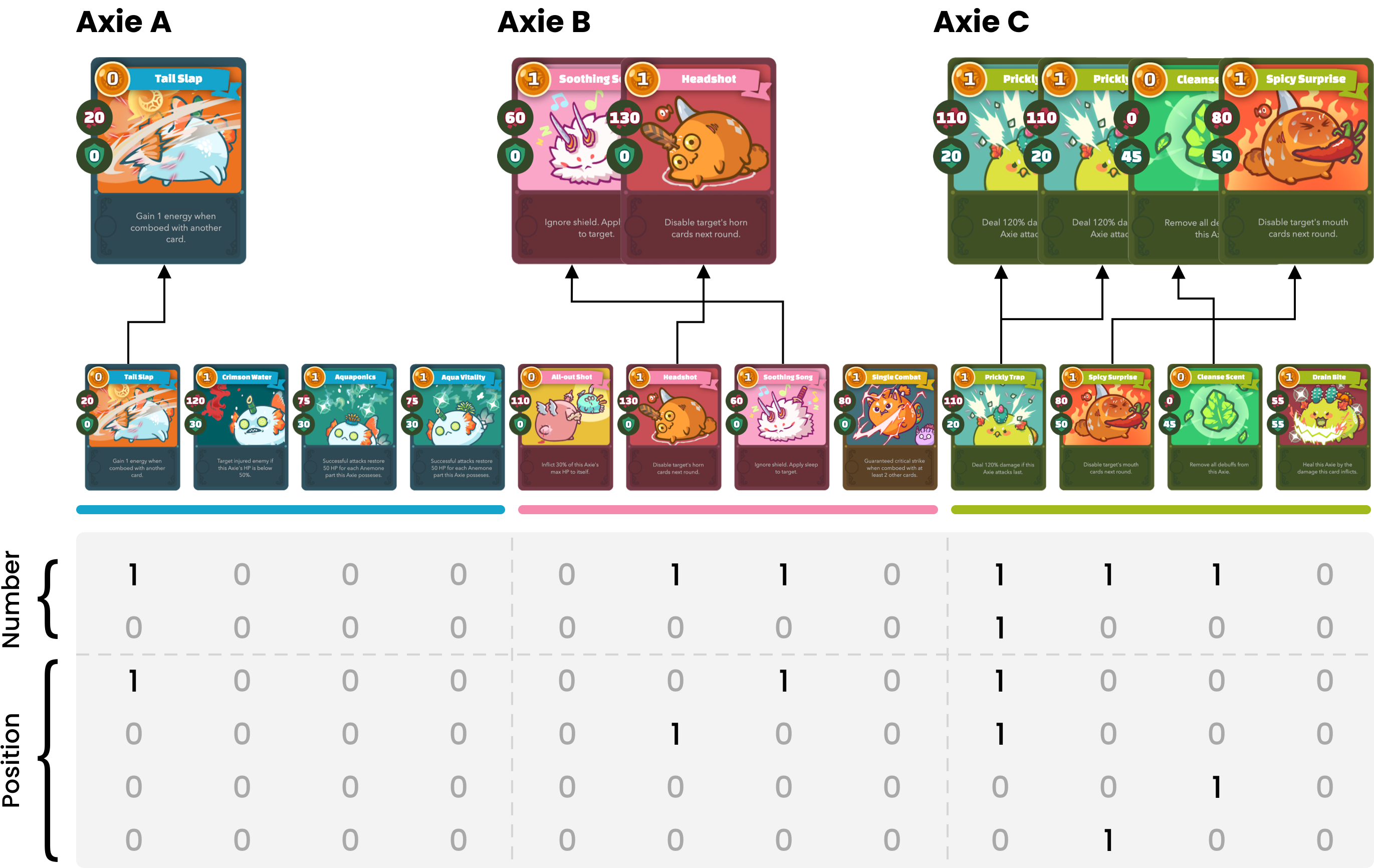}
    \caption{A demonstration of an encoded action consisting of 3 card sequences. Axie A/B/C respectively places 1/2/4 cards in this round. The status of each card is encoded in a 6-digit vector where the first two digits represent the number of this card and the rest 4 digits encode its positional information. The matrix is formed by 12 such vectors for 12 distinct cards.}
    \label{fig:action encoding}
    \end{figure}

In Axie Infinity, each team has three axies, each of which has two copies of 4 distinct cards. Thus, this forms a 24-card deck. In each round, the player places a sequence of cards for each axie, thus three sequences of cards are placed.    Each sequence can have at most 4 cards by the rule of the game.  Considering all these rules, we propose a novel action encoding scheme to vectorize an action as a $6\times12$ matrix. One example is given in Figure \ref{fig:action encoding}. All legit actions form the discrete action space $\mathcal{A}$ whose size equals 23,149,125. In contrast, DouDizhu in \cite{zha2021douzero} only has 27,472 actions.

% The state transition function is denoted as $P_t: \mathcal{S}\times \mathcal{A}\times{S}\rightarrow [0,1]$.

    % Axie Infinity is an episodic card game. Both players select the cards to play in each round. The card executions take place after both players confirm the chosen cards. 

    % (win/loss) or when the maximal step $T$ is reached (tie). % we could add some illustration on imperfect games

    % The history information about the cards which have been placed in the past are included to infer the distribution of the opponent's card. 
    
    % encoded using LSTM (cite), then the encoded history is concatenated with all other regular information and processed by multiple layers of MLP.
    % \item \textit{Action Space $\mathcal{A}$}\\

    % In the first round of a game, each player randomly acquires 5 (?) cards from the card deck. In every following round, each player randomly acquires 3 (?) cards from the rest of the card deck. 

    % The effect of these cards is highly sensitive to the card order. 

    % In every round, each player places a sequence of cards, which are constrained by a set of rules: 1)

    % we propose a new action encoding scheme to vectorize the sequence of cards to play in each round. 

    \vspace{-1em}
    % \item \textit{Reward}\\
    % Though reward shaping can effectively facilitate training, we would like to introduce minimal human knowledge to the model. Non-zero rewards are given at the end of each game. The game rule requires player to discard cards if the player holds more than 12 cards at the end of a round. As the player can not benefit from discarding cards, we penalize the agent by adjusting the reward with the number of discarded cards. 
    
    We define the reward as the result of the game, with a penalty on the activity discarding cards. Mathematically, we denote the terminal state set $\mathcal{T}$ as a set of states at which the game is over. For a transition tuple $(s_t, a_t, s_{t+1})$, $0\leq t<T-1$, we define the reward function as 
    \begin{equation*}
        r_t(s_t, a_t, s_{t+1}) = 
        \begin{cases}
        I - c\cdot n_d,& \text{if } s_{t+1}\in \mathcal{T}, \\
        0,& o.w.,
        \end{cases}
    \end{equation*}
    where $I$ and $n_d$ are components in $s_{t+1}$. The game result indicator $I$ equals to 1/0/-1 if the agent wins/ties/loses the game, and $n_d$ is the number of discarded cards in the whole game. The positive constant $c$ adjusts the importance of the penalty term.

% A general MDP formulation involves a discount factor $\gamma$. We exclude this term because we focus on maximizing the expected \textit{total} reward to solve this problem, rather than the \textit{discounted} reward.   
%	The ‘failure’ of planning and the ‘ill-posed’ route can be explained in two ways. Firstly, the planning module highly depends on the initialization of all states before iteration, thus improper initial states, which are irrational for a real driving mission, may lead to a sub-optimal result. Secondly, it is difficult to find a proper path when the obstacle vehicles at the target lane is driving with low speed or the cut-in position is too far from the ego vehicle.
	
\section{Methodology}

% % why not MCTS：
% Silver et al \cite{silver2016mastering}, proposed a general RL framework for Go game using Monte Carlo Tree Search (MCTS). MCTS learns directly from experience by extending the tree search. However, MCTS cannot solve our problem because we have a huge action space which makes the searching process and learn the system dynamic model becomes  extremely time-consuming.  MCTS is not applicable in practice for our problem because of the time limit on the decision in each round. As a result, we consider an actor-critic style reinforcement learning algorithm to learn the policy.
% % 
% As a result, we use a model-free RL method. Monte-Carlo (MC) methods are traditional reinforcement learning algorithms based on averaging sample returns. It learn from the experience without relaying prior knowledge of transition models. MC methods are designed for episodic tasks, where experiences can be divided into episodes and all the episodes eventually terminate. To optimize a policy $\pi$, every-visit MC can be used to estimate Q-table Q(s, a)
% by iteratively executing the following procedure. 

% DouZero \cite{zha2021douzero} applies a value-based method, which selects the action with the highest Q-value among all available actions. This is not applicable to our problem because of the large size of the action space in Axie Infinity. 
Due to the challenges brought by the large-scale action space, we consider to only evaluate a small group of actions which are filtered out from all feasible actions. We form this small set of actions with those actions which have similar effects with a target action. This target action is generated by a policy function. A distance function is needed to measure the ``similarity'' between two actions. 
% This is connected to the word embedding problem in Natural Language Process (NLP) field. 
Inspired by Word2Vec \cite{mikolov2013efficient}, we learn an action embedding function which maps one-hot-like action vectors  into a continuous space which is defined as the latent action space. The Euclidean distance in this latent space reflects the difference between two actions in terms of their effects. 

% We notice that our method is similar to \cite{dulac2015deep}. The key difference is that \cite{dulac2015deep} uses the Euclidean distance in the original action space instead of the latent action space.

% They select the candidates based on the Euclidean distance between Their policy function produces a target action, and forms a set with those actions which are ``close'' to this target action. The distance is defined as the Euclidean distance between two action vectors. However, this distance usually cannot tell how ``far'' two actions are away from each other. For instance, in one-hot encoded action space, the distances between an arbitrary action with any other actions are equal to a constant $\sqrt{2}$. Ideally, in our task, the distance should reflect how different one action is from another in terms of their effects. 
% For example, two cards of attack should have closer distance than that between a card of defence and a card of attack, but this method gives equal distances between. 

% We propose the following method to solve the MDP in \ref{sec:formulation}. 
% We present how our method selects actions and how to train the components of the policy function in the following sections. 

\begin{figure*}
\centering
\includegraphics[width=0.8\textwidth]{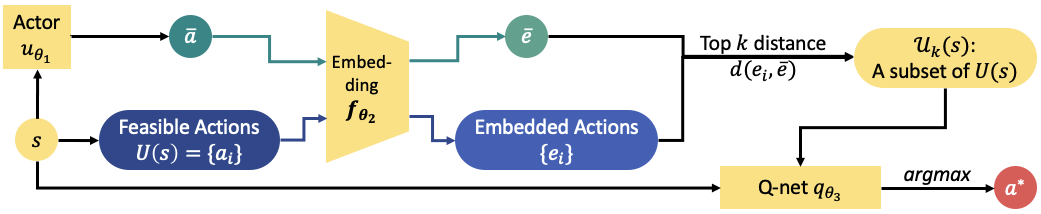}
\caption{An illustration on the decision procedure. }\label{fig:overall policy}
\end{figure*}
\subsection{
A Decision Procedure}
We illustrate the decision procedure in Figure \ref{fig:overall policy}.
This decision procedure of our method consists of 3 parametrized components: 
\begin{enumerate}
    \item A raw policy function  $u_{\theta_1}: \mathcal{S}\rightarrow\mathbb{R}^n$ where $n$ is the dimensions of the action space, $\mathcal{A}\subset \mathbb{R}^n$.
    \item An embedding function $f_{\theta_2}: \mathbb{R}^n\rightarrow\mathcal{E}$ where $\mathcal{E}$ is the latent action space, and $\mathcal{E}\subset \mathbb{R}^m$, $m<n$. For an arbitrary action $a$, $e = f_{\theta_2}(a)$ is called the latent action representation of $a$.
    \item A state-action value function, i.e., a Q-function  $q_{\theta_3}:\mathcal{S}\times\mathcal{A}\rightarrow\mathbb{R}$. The Q-value $q_{\theta_3}(s, a)$ evaluates the expected return when executing action $a$ at state $s$. 
\end{enumerate}
% This method is inspired by \cite{chandak2019learning} and \cite{dulac2015deep}. We highlight how our method differs from their works after 
% To explain how this method perform controls, we denote the following components: 
% $f_{\theta_1}: $

For a given input $s$, the overall policy function $\mu$ selects the action using the following procedure. 
Firstly, we obtain a point $\Bar{a} = u_{\theta_1}(s)$ as a ``raw action'', note it is possible that $\Bar{a}\not\in\mathcal{A}$.
Secondly, we denote the set of all available actions for $s$ as $U(s)$. 
% We obtain a latent representation of the raw action $\Bar{e}=f_{\theta_2}(\Bar{a})$ and the set of latent representations of the available actions $\{f_{\theta_2}(a); a\in U(s)\}$. Then, 
We calculate the distance between available actions with the raw action in the latent space by $d(a, \Bar{a}; f_{\theta_2})=\|f_{\theta_2}(a)-f_{\theta_2}(\Bar{a})\|^2$ for all $a\in U(s)$. We form a $k$-element subset of $U(s)$ with the top $k$ closest actions to the raw action in the latent space, denote it as $\mathcal{U}_k(s;\theta_1, \theta_2)$. Mathematically, this is done by
\begin{equation}
    \mathcal{U}_k(s;\theta_1, \theta_2) = \argmin_{\mathcal{U}\subset U(s), |\mathcal{U}|=k} \sum_{a\in \mathcal{U}}d(a, u_{\theta_1}(s); f_{\theta_2}).
\end{equation}
In the last, we select the action which has the highest Q-value in this subset. If we denote an overall policy function as $\mu$ for a given state $s$, the action is selected by
\begin{equation}\label{eq:overall policy}
    a^* = \mu(s; \theta_1, \theta_2, \theta_3) = \argmax_{a\in\mathcal{U}_k(s;\theta_1, \theta_2)}q_{\theta_3}(s, a).
\end{equation}

% \lipsum[1-2]

% \lipsum[3-10]

% \begin{figure}[h]
% \centering
% \includegraphics[width=10cm]{img/learningframework2.png}
% \caption{learning framework.}
% \label{tripdur}
% \end{figure}
\subsection{Training procedures}
Our method consists of three sets of parameters $\theta_1$, $\theta_2$, and $\theta_3$. We design a two-stage algorithm to train these parameters.
% In the first stage, we apply a supervised learning procedure to learn the embedding function $f_{\theta_2}$. In the second stage, we train the policy function and Q-function $\theta_1$ and $\theta_3$ using RL procedure that is similar to DDPG \cite{lillicrap2015continuous}. 

\begin{figure}
\centering
\includegraphics[width=0.35\textwidth]{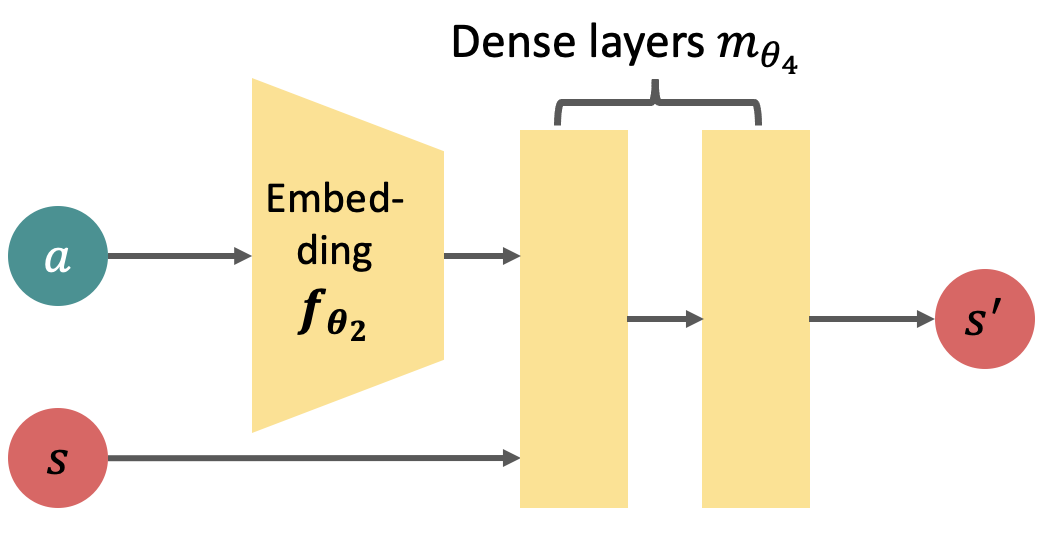}
\caption{An illustration of the model of the system dynamics.}
\label{fig:system_model}
\end{figure}

In the first stage of training, we design a supervised learning method to learn the embedding function $f_{\theta_2}$.
We learn the action representations based on the effects of the actions on the system states. For instance, assume an arbitrary state $s\in\mathcal{S}$ and two actions $a_1$ and $a_2$, if the probability measure $p(s, a_1)$ is similar to $p(s, a_2)$ where $p(s, a) = P(S_{t+1}\mid S_t = s, A_t = a)$, we say $a_1$ and $a_2$ have similar effects. Following this idea, we train the embedding function by learning a model of the system dynamics which consists of the embedding function $f_{\theta_2}$. We illustrate the architecture of the system model in Figure \ref{fig:system_model}. Specifically, we define a deterministic transition function $m_{\theta_4}: \mathcal{S}\times\mathcal{\mathcal{E}}\rightarrow\mathcal{S}$, which maps the state and action embedding to the next state. Thus, $m_{\theta_4}(s, f_{\theta_2}(a))$ should estimate the next state after executing $a$ at $s$.  We define the following objective function to minimize the mean square error between the estimated next states with the actual next states
\begin{equation}\label{eq:pretrain}
    J_1(\theta_2, \theta_4) = \mathbb{E}_{P_t, \cdot}[(m_{\theta_4}(s, f_{\theta_2}(a))-s')^2].
\end{equation}
We collect the transition tuples $(s, a, s')$ in a dataset $D$ by randomly selecting actions. Then, we apply a gradient-based optimization method to optimize $\theta_2$ and $\theta_4$ by minimizing $J_1$ on $D$. In the next stage of training, we discard the dense layers, and only use the embedding function $\theta_2$.

In the second stage of training, similar to Deep Deterministic Policy Gradient in \cite{lillicrap2015continuous},
% similar to Policy Iteration (PI) \cite{bertsekas2011dynamic}, 
we apply an iterative training procedure to alternatively update the deterministic raw policy function (the actor) $u_{\theta_1}$ and the Q-function (the critic) $q_{\theta_3}$. We use the raw action $\Bar{a}$ instead of the final action $a^*$ in the policy improvement part \cite{dulac2015deep}. The Q-function training uses Monte-Carlo estimate described in \cite{sutton2018reinforcement}.
% For $u_{\theta_1}$, we improve the policy by
% \begin{equation}\label{eq:policy}
%     J_2(\theta_1) = \mathbb{E}_{\tau\sim P_t, u_{\theta_1}}[\sum_{t=0}^Tq_{\theta_3}(s_t, a_t)].
% \end{equation}
% In terms of $q_{\theta_3}$, we apply Monte-Carlo estimate method in \cite{sutton2018reinforcement} to calibrate $q_{\theta_3}$,
% \begin{equation}\label{eq:value}
%     J_3(\theta_3) = \mathbb{E}_{\tau\sim P_t, u_{\theta_1}}\left[\sum_{t=0}^{T-1}(q_{\theta_3}(s_t, a_t) - \sum_{i = t}^T r_i(s_i, a_i, s_{i+1}))^2\right]
% \end{equation}

% We apply an iterative updating procedure to train $\theta_1$ and $\theta_3$, as described in Algorithm \ref{alg:training}.

\section{Experiment}
% ID to test: 149， 130， 199， 148， 186， 140

% dmc baseline, 运算时间
In this section, we compare our method with two baseline methods:
\begin{enumerate}
    \item DouZero. We adapt the DouZero method in \cite{zha2021douzero} to our problem. We design a similar action encoding scheme as the one mentioned in this paper, where each action is encoded as a 2-by-12 matrix. 
    % Similar to action encoding method This is a value-based RL method that uses Monte-Carlo estimate to calibrate the Q-function. 
    \item DouZero+pooling. We reduce the size of the action space by shrinking its dimensions. We design this baseline method by adding an 1D pooling layer \cite{yu2014mixed} on the flattened actions from DouZero. 
\end{enumerate}
Indeed, other techniques to reduce the scale of action spaces are mentioned in \cite{kanervisto2020action}. They inevitably introduce prior human knowledge, which conflicts with our intention, and this makes the comparison unfair. In our experiments, we try to answer the following questions:
\begin{enumerate}
    \item Does our method produces overall better performance than other baseline methods?
    \item Does our method achieve better sample efficiency?
\end{enumerate}

\subsection{Sample Efficiency}
We select six teams that are popular at different levels in the global rank. A detailed description of these teams can be found in Appendix \ref{app:axie}. We train a model for each team using 3 methods: our method, the DouZero method, and the DouZero+pooling method. Thus, we trained 18 models in total. To make a fair comparison, we stop the training after $1\times 10^7$ steps for all these models. 

\begin{figure}
\centering
\includegraphics[width=8cm]{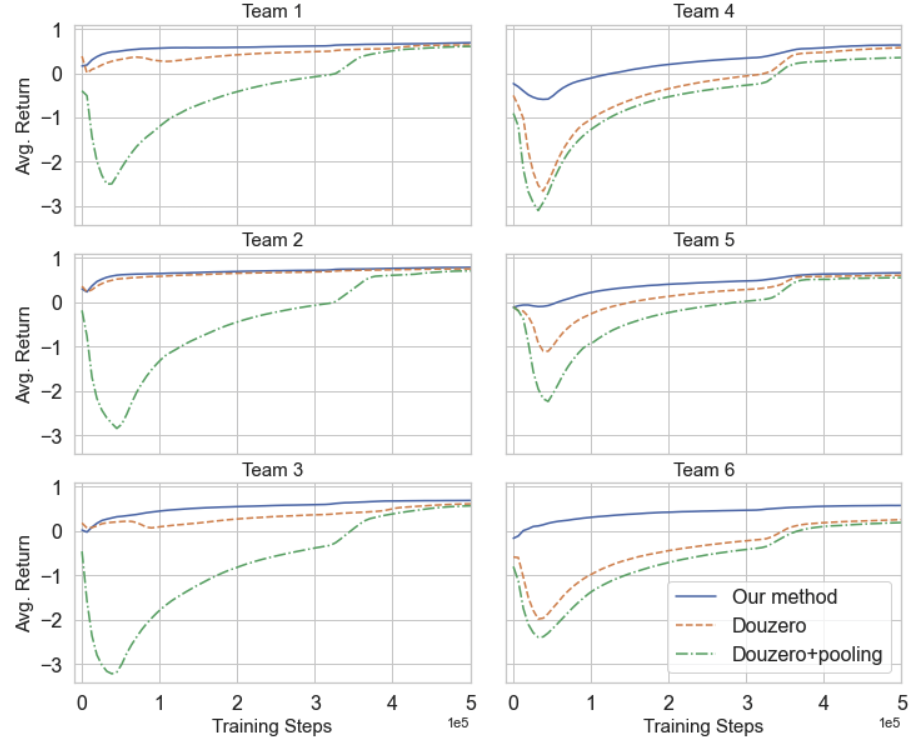}
\caption{\textbf{Comparison on the learning curves.} Six line-charts shows the learning curves of 18 models in the first $5\times 10^5$ steps. It can be seen that our method shows a faster convergence with the same amount of samples on most of the 6 teams. }
\label{frame}
\end{figure}

Figure \ref{frame} shows the evolution of average returns during training in the first $1\times 10^5$ steps. It can be seen that, on most of the teams, the models from our method have the highest average return among the three by trained with the same amount of samples. This indicates that our method can quickly produce high-quality models due to high sample efficiency. 

\subsection{Battle}
We evaluate the performance of each model by letting it play against other models and rule-based players. This guarantees that the models trained for the same team have the same set of opponents. Each battle consists of 1000 games. Each model plays 29000 games against various opponents to obtain a comprehensive result. This makes the variance of estimators of the winning rates small enough to show statistical significance. 

We aggregate the winning rate of the models trained by each method in Table \ref{winningrate}. It can be seen that the overall winning rate of our method is 5\% and 7\% higher than the DouZero and DouZero+pooling method respectively. This also confirms our method has a better generalization ability on different teams.  
\vspace{-0.5em}
\begin{table}[H]
	\centering
	\caption{The average wining rate of 3 methods on 6 teams \\(174000 games for each method)}\label{winningrate}
	\begin{adjustbox}{max width=0.45\textwidth}
	\begin{tabular}{cccc}\hline\hline \noalign{\smallskip}
		Method & Winning rate  \\
		\noalign{\smallskip}\hline\noalign{\smallskip}
		Our method & \textbf{0.4986} \\ %($\pm$ 0.2090)} \\
		\noalign{\smallskip}\hline\noalign{\smallskip}
		DouZero  &  0.4477 \\ %($\pm$ 0.2361) \\
		\noalign{\smallskip}\hline\noalign{\smallskip}
		DouZero+pooling & 0.4283 \\ %($\pm$ 0.2148)   \\

		\noalign{\smallskip}\hline\hline
	\end{tabular}
	\end{adjustbox}
	\label{neighbor}

\end{table}

\begin{figure}
\centering
\includegraphics[width=0.5\textwidth]{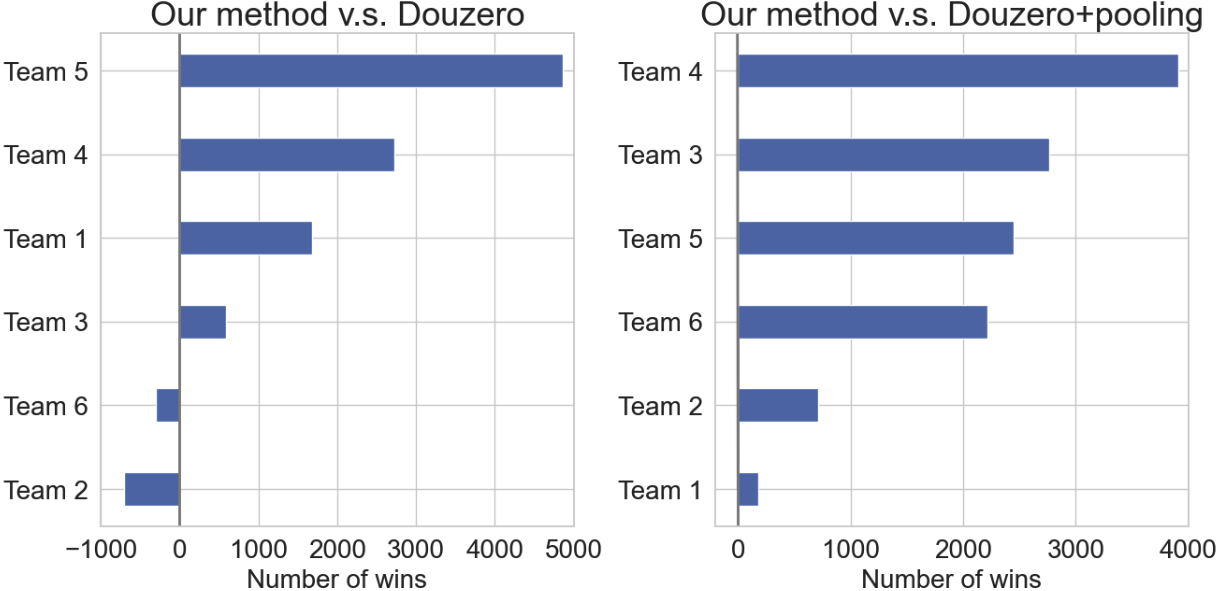}
\caption{\textbf{Comparison on battle performance between our method with two baseline methods on 6 teams.} We train 18 models using these three methods on 6 teams. We evaluate each model by letting it play against a same set of opponent players consisting of the other trained agents and random players (29000 battles in total).  Note the qualities in the left/right bar-plot respectively show (the number of wins of our method - the number of wins of the DouZero/DouZero+pooling method). }
\label{battle}
\vspace{-1em}
\end{figure}

We also compare the models which use the same team. For each team, we calculate the difference between the number of wins by our method with that of DouZero or DouZero+pooling. A positive difference indicates our method plays this team better than the baseline method.  We visualize these differences in the number of wins in Figure~\ref{battle}. We can find that our method can achieve a higher number of wins across most teams than the baseline methods. This indicates that our method has good generalizability to most team types. We notice that our method cannot outperform DouZero on teams 2 and 6. The reason is that strategies for these two teams are more diversified than those for the other teams. This makes high-reward actions far away from each other in the latent space. In such cases, the DouZero method may make better decisions than our method because it evaluates every feasible actions, though it is more computationally expensive.

\subsection{Time of Action Selections}
Axie Infinity requires players to make decisions in 30 seconds in every round. We mention that some methods such as Monte Carlo Tree Search (MCTS) in \cite{silver2018general} can also achieve good performance, but they usually fail to select cards by the time limit. We show our method is more efficient in time than the other two baseline methods using the time consumed by selecting card sequences. We test the 18 models in 1000 rounds, and record the time for selecting cards\footnote{The experiment is conducted on a platform with Apple M1 chip (3.2 GHz) and 16 GB LPDDR4 RAM.}. Table \ref{time} shows the statistics of the distribution of time for card selections for three methods. It can be seen that our method uses shorter time than the other two methods in average. The 25\% quantiles for the three methods are similar, but the 75\% quantiles of the two baseline methods is mush higher than that of our method. This indicates that our method has more consistent performance in terms of the time. This confirms that our method benefits from only evaluating actions in a fixed-size subset from all feasible actions. 
% mean	7.423614	18.713344	16.352037
% std	18.422932	86.675792	72.285659
% min	2.054930	0.889063	0.891209
% 25%	5.012751	4.917085	5.223632
% 50%	5.648136	8.718610	8.608937
% 75%	7.001162	16.502500	15.872359
% max	1186.318159	3699.200869	3707.821131
\begin{table}[H]
	\centering
	\caption{The distribution of time for selecting cards by three methods (in millisecond)}\label{time}
	\begin{adjustbox}{max width=0.45\textwidth}
	\begin{tabular}{cccc}\hline\hline \noalign{\smallskip}
		Statistics & Our Method & DouZero & DouZero+pooling  \\
		\noalign{\smallskip}\hline\noalign{\smallskip}
		Mean & \textbf{7.42} & 18.71 &  16.35  \\ %($\pm$ 0.2090)} \\
		\noalign{\smallskip}\hline\noalign{\smallskip}
		Std  &  \textbf{18.42}&86.67 &72.28 \\ %($\pm$ 0.2361) \\
		\noalign{\smallskip}\hline\noalign{\smallskip}
		25\% Quantile & 5.01 &4.91 &5.22 \\ %($\pm$ 0.2148)   \\
		\noalign{\smallskip}\hline\noalign{\smallskip}
        50\% Quantile & \textbf{5.64} & 8.71&8.61\\
        \noalign{\smallskip}\hline\noalign{\smallskip}
        75\% Quantile &\textbf{7.00} &16.50 & 15.87\\
        \noalign{\smallskip}\hline\noalign{\smallskip}
        Max & \textbf{1186.32} & 3699.20 & 3707.82 \\
		\noalign{\smallskip}\hline\hline
	\end{tabular}
	\end{adjustbox}
	\label{neighbor}

\end{table}

\section{Conclusions}
In this study, we try to use an RL method to solve the challenges in a card game problem with a large action space. We give the MDP formulation for the game Axie Infinity. We propose a general RL algorithm to learn the strategies of different teams. We design a training procedure to learn the action embedding function without prior information. We empirically demonstrate our method outperforms the baseline methods in terms of the battle performance and the sample efficiency. 

Our work can be improved by incorporating the self-play technique \cite{silver2018general} in training to enhance the opponents.
Moreover, learning prior information from the textual data of card descriptions can enhance the action embedding component in our method. 
Future works may focus on this direction for further improvement. 
% \section{Discussion}

% \section{ Acknowledgments}
	
	\bibliographystyle{IEEEtran}
	\bibliography{arxiv} 

% Generated by IEEEtran.bst, version: 1.14 (2015/08/26)
\begin{thebibliography}{10}
\providecommand{\url}[1]{#1}
\csname url@samestyle\endcsname
\providecommand{\newblock}{\relax}
\providecommand{\bibinfo}[2]{#2}
\providecommand{\BIBentrySTDinterwordspacing}{\spaceskip=0pt\relax}
\providecommand{\BIBentryALTinterwordstretchfactor}{4}
\providecommand{\BIBentryALTinterwordspacing}{\spaceskip=\fontdimen2\font plus
\BIBentryALTinterwordstretchfactor\fontdimen3\font minus
  \fontdimen4\font\relax}
\providecommand{\BIBforeignlanguage}[2]{{%
\expandafter\ifx\csname l@#1\endcsname\relax
\typeout{** WARNING: IEEEtran.bst: No hyphenation pattern has been}%
\typeout{** loaded for the language `#1'. Using the pattern for}%
\typeout{** the default language instead.}%
\else
\language=\csname l@#1\endcsname
\fi
#2}}
\providecommand{\BIBdecl}{\relax}
\BIBdecl

\bibitem{mnih2015human}
V.~Mnih, K.~Kavukcuoglu, D.~Silver, A.~A. Rusu, J.~Veness, M.~G. Bellemare,
  A.~Graves, M.~Riedmiller, A.~K. Fidjeland, G.~Ostrovski \emph{et~al.},
  ``Human-level control through deep reinforcement learning,'' \emph{nature},
  vol. 518, no. 7540, pp. 529--533, 2015.

\bibitem{lillicrap2015continuous}
T.~P. Lillicrap, J.~J. Hunt, A.~Pritzel, N.~Heess, T.~Erez, Y.~Tassa,
  D.~Silver, and D.~Wierstra, ``Continuous control with deep reinforcement
  learning,'' \emph{arXiv preprint arXiv:1509.02971}, 2015.

\bibitem{mnih2016asynchronous}
V.~Mnih, A.~P. Badia, M.~Mirza, A.~Graves, T.~Lillicrap, T.~Harley, D.~Silver,
  and K.~Kavukcuoglu, ``Asynchronous methods for deep reinforcement learning,''
  in \emph{International conference on machine learning}.\hskip 1em plus 0.5em
  minus 0.4em\relax PMLR, 2016, pp. 1928--1937.

\bibitem{schulman2017proximal}
J.~Schulman, F.~Wolski, P.~Dhariwal, A.~Radford, and O.~Klimov, ``Proximal
  policy optimization algorithms,'' \emph{arXiv preprint arXiv:1707.06347},
  2017.

\bibitem{zha2021douzero}
D.~Zha, J.~Xie, W.~Ma, S.~Zhang, X.~Lian, X.~Hu, and J.~Liu, ``Douzero:
  Mastering doudizhu with self-play deep reinforcement learning,'' in
  \emph{International Conference on Machine Learning}.\hskip 1em plus 0.5em
  minus 0.4em\relax PMLR, 2021, pp. 12\,333--12\,344.

\bibitem{bowling2015heads}
M.~Bowling, N.~Burch, M.~Johanson, and O.~Tammelin, ``Heads-up limit hold’em
  poker is solved,'' \emph{Science}, vol. 347, no. 6218, pp. 145--149, 2015.

\bibitem{brown2018superhuman}
N.~Brown and T.~Sandholm, ``Superhuman ai for heads-up no-limit poker: Libratus
  beats top professionals,'' \emph{Science}, vol. 359, no. 6374, pp. 418--424,
  2018.

\bibitem{brown2019superhuman}
------, ``Superhuman ai for multiplayer poker,'' \emph{Science}, vol. 365, no.
  6456, pp. 885--890, 2019.

\bibitem{li2020suphx}
J.~Li, S.~Koyamada, Q.~Ye, G.~Liu, C.~Wang, R.~Yang, L.~Zhao, T.~Qin, T.-Y.
  Liu, and H.-W. Hon, ``Suphx: Mastering mahjong with deep reinforcement
  learning,'' \emph{arXiv preprint arXiv:2003.13590}, 2020.

\bibitem{guan2022perfectdou}
Y.~Guan, M.~Liu, W.~Hong, W.~Zhang, F.~Fang, G.~Zeng, and Y.~Lin, ``Perfectdou:
  Dominating doudizhu with perfect information distillation,'' \emph{arXiv
  preprint arXiv:2203.16406}, 2022.

\bibitem{dulac2015deep}
G.~Dulac-Arnold, R.~Evans, H.~van Hasselt, P.~Sunehag, T.~Lillicrap, J.~Hunt,
  T.~Mann, T.~Weber, T.~Degris, and B.~Coppin, ``Deep reinforcement learning in
  large discrete action spaces,'' \emph{arXiv preprint arXiv:1512.07679}, 2015.

\bibitem{chandak2019learning}
Y.~Chandak, G.~Theocharous, J.~Kostas, S.~Jordan, and P.~Thomas, ``Learning
  action representations for reinforcement learning,'' in \emph{International
  conference on machine learning}.\hskip 1em plus 0.5em minus 0.4em\relax PMLR,
  2019, pp. 941--950.

\bibitem{kanervisto2020action}
A.~Kanervisto, C.~Scheller, and V.~Hautam{\"a}ki, ``Action space shaping in
  deep reinforcement learning,'' in \emph{2020 IEEE Conference on Games
  (CoG)}.\hskip 1em plus 0.5em minus 0.4em\relax IEEE, 2020, pp. 479--486.

\bibitem{mikolov2013efficient}
T.~Mikolov, K.~Chen, G.~Corrado, and J.~Dean, ``Efficient estimation of word
  representations in vector space,'' \emph{arXiv preprint arXiv:1301.3781},
  2013.

\bibitem{sutton2018reinforcement}
R.~S. Sutton and A.~G. Barto, \emph{Reinforcement learning: An
  introduction}.\hskip 1em plus 0.5em minus 0.4em\relax MIT press, 2018.

\bibitem{yu2014mixed}
D.~Yu, H.~Wang, P.~Chen, and Z.~Wei, ``Mixed pooling for convolutional neural
  networks,'' in \emph{International conference on rough sets and knowledge
  technology}.\hskip 1em plus 0.5em minus 0.4em\relax Springer, 2014, pp.
  364--375.

\bibitem{silver2018general}
D.~Silver, T.~Hubert, J.~Schrittwieser, I.~Antonoglou, M.~Lai, A.~Guez,
  M.~Lanctot, L.~Sifre, D.~Kumaran, T.~Graepel \emph{et~al.}, ``A general
  reinforcement learning algorithm that masters chess, shogi, and go through
  self-play,'' \emph{Science}, vol. 362, no. 6419, pp. 1140--1144, 2018.

\end{thebibliography}

\section{Appendix}
\subsection{Game rules}
We briefly explain the rules of Axie Infinity in this section. More detailed rules can be found in the website (\url{https://whitepaper.axieinfinity.com/}).

\begin{figure}
\centering
\includegraphics[width=8cm]{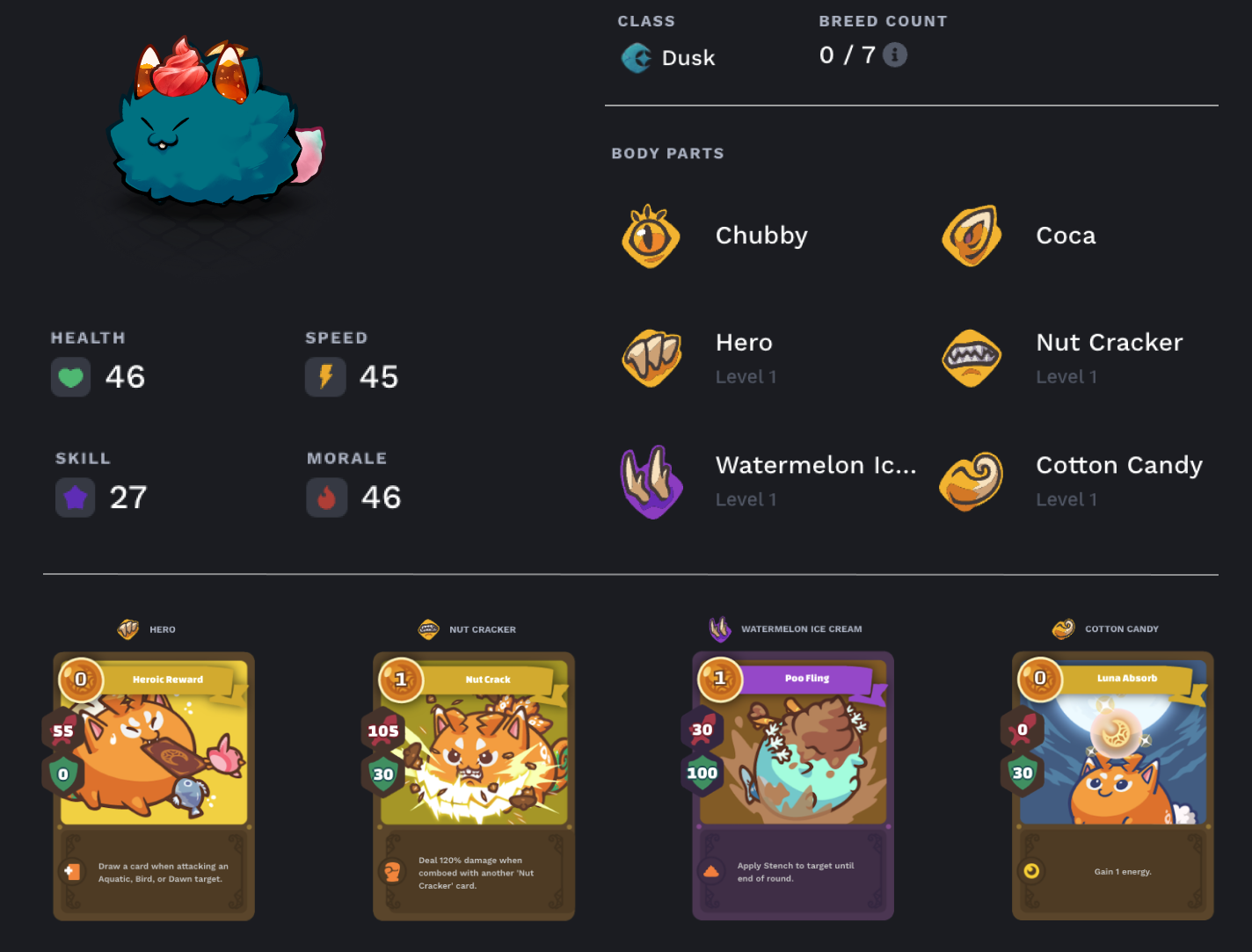}
\caption{An illustration of the attributes of an Axie.}
\label{axie_attribute}
\end{figure}

\begin{figure}
\centering
\includegraphics[width=8cm]{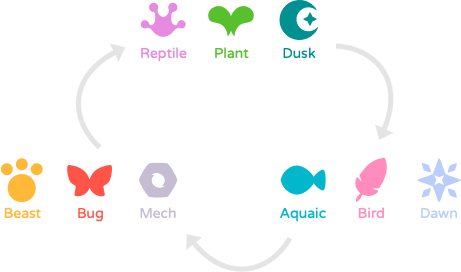}
\caption{The relationship between the 9 classes.}
\label{axie_class}
\end{figure}

Axie Infinity is an online one-verse-one card game. Axies are virtual pets in this game, and they can be traded between players. Figure \ref{axie_attribute} illustrates the attributes of an Axie. The behavior of an Axie can be determined by three of its characteristics: the class, the attributes (stats), and the cards. The game has 9 classes in total. Every class is stronger than three classes, and is weaker to other three classes. Figure \ref{axie_class} depicts the relationship between 9 classes.  In battles, an Axie produces more/less damages to other Axies if their class are weak/strong against the class of this Axie. Attributes (stats) consist of 4 values which indicate the Axie's ability in Health/Speed/Skill/Morale. Every Axie carries 8 cards -- 2 copies of 4 distinct cards.
Players need to buy three Axies to form a team for combat. Thus, this forms a 24-card deck for the player to play. 

\begin{figure}
\centering
\includegraphics[width=8cm]{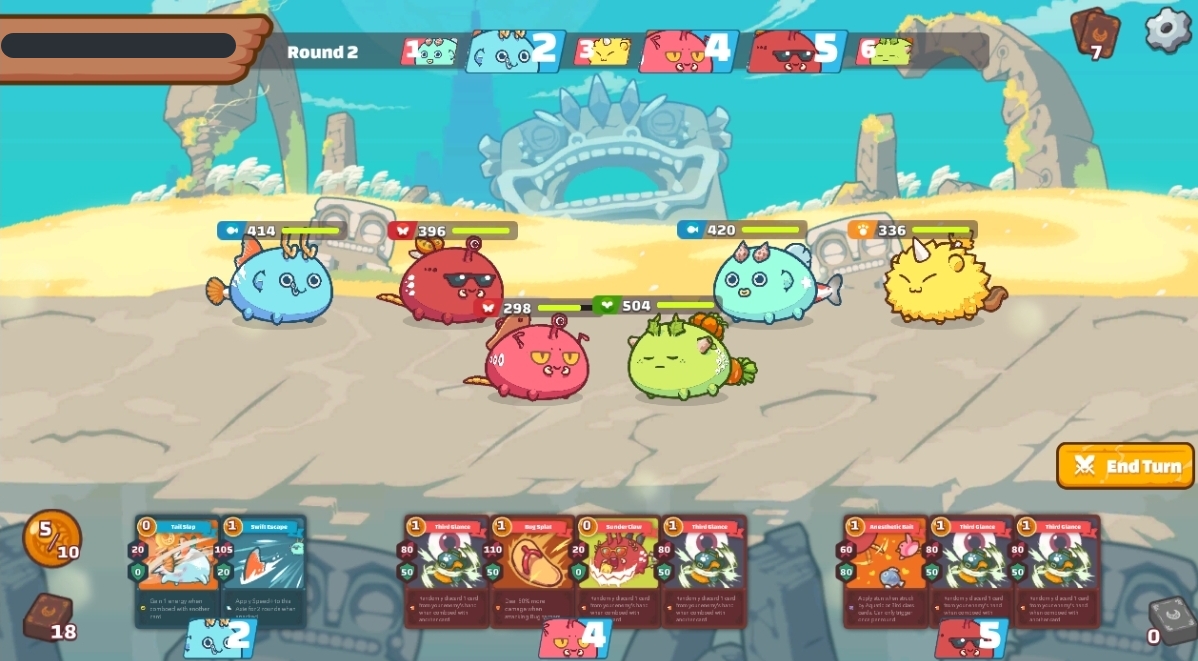}
\caption{A snapshot of a battle of Axie Infinity.}
\label{axie_battle}
\end{figure}

Figure \ref{axie_battle} demonstrates a battle between two players. Two players play cards round by round. In each round, the player draws cards from the card deck and gains energy points. The energy points limit the number of cards to select since some cards need energy points to execute. Most of the cards consume 0 or 1 energy point. The player can select at most 4 cards from an Axie, the order of these 4 cards can significantly impact their effects. Thus, the player create 3 card sequences from each Axie to play in a round. 
Once the card sequences are determined, the player can hit ``End Turn'' button, and the card sequences submitted by the two players are executed in an order which is associated with the Axies' attributes and the card effects. The cards produce damages to the Axies on the other side. Each Axie has health points, the Axie is beaten when it has 0 health point. When all three Axies on a side are beaten, the game ends, and the player whose at least one Axie is alive wins this game. If all six Axies are beaten, a tie is reached. If the game can continue after the card execution, next round starts and the players draw cards from the remaining card deck again. Once the card deck is exhausted, the used cards are automatically shuffled and form a new card deck.

\subsection{Teams in the experiments}\label{app:axie}
% [3, 130, 140, 149, 169, 199]
% Stats: Health | Speed | Skill | Morale

In order to validate our method, we select six teams which are popular from different levels in the global rank. Since the human strategies for these six teams are very different, our method ideally can lean these diversified strategies. The following tables show the detailed information for the 6 teams in our experiments.

\begin{table}[H]\
	\centering
	\caption{Details of the Six teams in the experiments}
	\begin{adjustbox}{max width=0.6\textwidth}
	\begin{tabular}{cccc}\hline\hline \noalign{\smallskip}
		\textbf{Team 1} & Axie 1 & Axie 2 & Axie 3  \\
		\noalign{\smallskip}\hline\noalign{\smallskip}
		Axie Class & \text Aquatic & Plant & Bird  \\ %($\pm$ 0.2090)} \\
		\noalign{\smallskip}\hline\noalign{\smallskip}
		Axie Stats & \text 45/57/35/27 & 61/31/31/41 & 27/59/35/43 \\ %($\pm$ 0.2361)\\
		\noalign{\smallskip}\hline\noalign{\smallskip}
		Axie Card 1 & Aqua Vitality & Cleanse Scent & Single Combat \\ %($\pm$ 0.2148)   \\
		\noalign{\smallskip}\hline\noalign{\smallskip}
		Axie Card 2 & Crimson Water & Drain Bite & Soothing Song \\ %($\pm$ 0.2148)   \\
		\noalign{\smallskip}\hline\noalign{\smallskip}
		Axie Card 3 & Aquaponics & Prickly Trap & Headshot \\ %($\pm$ 0.2148)   \\
		\noalign{\smallskip}\hline\noalign{\smallskip}
		Axie Card 4 & Tail Slap & Spicy Surprise & All-out Shot \\ %($\pm$ 0.2148)   \\

		\noalign{\smallskip}\hline\hline
	\end{tabular}
	\end{adjustbox}

\end{table}
% 3
% Axie: Aquatic
% Cards: Aqua Vitality | Crimson Water | Aquaponics | Tail Slap
% Stats: 45 | 57 | 35 | 27 
% Axie: Plant
% Cards: Cleanse Scent | Drain Bite | Prickly Trap | Spicy Surprise
% Stats: 61 | 31 | 31 | 41
% Axie: Bird
% Cards: Single Combat | Soothing Song | Headshot | All-out Shot 
% Stats: 27 | 59 | 35 | 43
\begin{table}[H]
	\centering

	\begin{adjustbox}{max width=0.6\textwidth}
	\begin{tabular}{cccc}\hline\hline \noalign{\smallskip}
		\textbf{Team 2} & Axie 1 & Axie 2 & Axie 3  \\
		\noalign{\smallskip}\hline\noalign{\smallskip}
		Axie Class & \text Aquatic & Bug & Dusk  \\ %($\pm$ 0.2090)} \\
		\noalign{\smallskip}\hline\noalign{\smallskip}
		Axie Stats & \text 45/51/35/33 & 43/35/35/51 & 57/41/27/39 \\ %($\pm$ 0.2361)\\
		\noalign{\smallskip}\hline\noalign{\smallskip}
		Axie Card 1 & Swift Escape & Scarab Curse & Ivory Chop \\ %($\pm$ 0.2148)   \\
		\noalign{\smallskip}\hline\noalign{\smallskip}
		Axie Card 2 & Terror Chomp & Terror Chomp & Terror Chomp \\ %($\pm$ 0.2148)   \\
		\noalign{\smallskip}\hline\noalign{\smallskip}
		Axie Card 3 & Bug Signal & Bug Signal & Bug Signal \\ %($\pm$ 0.2148)   \\
		\noalign{\smallskip}\hline\noalign{\smallskip}
		Axie Card 4 & Tail Slap & Anesthetic Bait & Cattail Slap \\ %($\pm$ 0.2148)   \\

		\noalign{\smallskip}\hline\hline
	\end{tabular}
	\end{adjustbox}

\end{table}
% 130
% Axie: Aquatic
% Cards: Swift Escape | Terror Chomp | Bug Signal | Tail Slap
% Stats: 45 | 51 | 35 | 33
% Axie: Bug
% Cards: Scarab Curse | Terror Chomp | Bug Signal | Anesthetic Bait
% Stats: 43 | 35 | 35 | 51
% Axie: Dusk 
% Cards: Ivory Chop | Terror Chomp | Bug Signal | Cattail Slap
% Stats: 57 | 41 | 27 | 39

\begin{table}[H]
	\centering
	\begin{adjustbox}{max width=0.6\textwidth}
	\begin{tabular}{cccc}\hline\hline \noalign{\smallskip}
		\textbf{Team 3} & Axie 1 & Axie 2 & Axie 3  \\
		\noalign{\smallskip}\hline\noalign{\smallskip}
		Axie Class & \text Plant & Dusk & Mech  \\ %($\pm$ 0.2090)} \\
		\noalign{\smallskip}\hline\noalign{\smallskip}
		Axie Stats & \text 56/33/31/44 & 59/47/27/31 & 37/48/43/36 \\ %($\pm$ 0.2361)\\
		\noalign{\smallskip}\hline\noalign{\smallskip}
		Axie Card 1 & October Treat & Ivory Chop & Juggling Balls \\ %($\pm$ 0.2148)   \\
		\noalign{\smallskip}\hline\noalign{\smallskip}
		Axie Card 2 & Vegetal Bite & Sneaky Raid & Sneaky Raid \\ %($\pm$ 0.2148)   \\
		\noalign{\smallskip}\hline\noalign{\smallskip}
		Axie Card 3 & Disguise & Surprise Invasion & Sinister Strike \\ %($\pm$ 0.2148)   \\
		\noalign{\smallskip}\hline\noalign{\smallskip}
		Axie Card 4 & Gas Unleash & Venom Spray & Twin Needle \\ %($\pm$ 0.2148)   \\

		\noalign{\smallskip}\hline\hline
	\end{tabular}
	\end{adjustbox}

\end{table}
% 140
% Axie: Plant
% Cards: October Treat | Vegetal Bite | Disguise | Gas Unleash
% Stats: 56 | 33 | 31 | 44
% Axie: Dusk
% Cards: Ivory Chop | Sneaky Raid | Surprise Invasion | Venom Spray
% Stats: 59 | 47 | 27 | 31
% Axie: Mech
% Cards: Juggling Balls | Sneaky Raid | Sinister Strike | Twin Needle
% Stats: 37 | 48 | 43 | 36

\begin{table}[H]
	\centering

	\begin{adjustbox}{max width=0.6\textwidth}
	\begin{tabular}{cccc}\hline\hline \noalign{\smallskip}
		\textbf{Team 4} & Axie 1 & Axie 2 & Axie 3  \\
		\noalign{\smallskip}\hline\noalign{\smallskip}
		Axie Class & \text Plant & Dusk & Dusk  \\ %($\pm$ 0.2090)} \\
		\noalign{\smallskip}\hline\noalign{\smallskip}
		Axie Stats & \text 59/31/31/43 & 57/43/27/37 & 51/50/27/36 \\ %($\pm$ 0.2361)\\
		\noalign{\smallskip}\hline\noalign{\smallskip}
		Axie Card 1 & October Treat & Barb Strike & Barb Strike \\ %($\pm$ 0.2148)   \\
		\noalign{\smallskip}\hline\noalign{\smallskip}
		Axie Card 2 & Vegetal Bite & Sneaky Raid & Chomp \\ %($\pm$ 0.2148)   \\
		\noalign{\smallskip}\hline\noalign{\smallskip}
		Axie Card 3 & Disguise & Surprise Invasion & Smart Shot \\ %($\pm$ 0.2148)   \\
		\noalign{\smallskip}\hline\noalign{\smallskip}
		Axie Card 4 & Gas Unleash & Venom Spray & Venom Spray \\ %($\pm$ 0.2148)   \\

		\noalign{\smallskip}\hline\hline
	\end{tabular}
	\end{adjustbox}

\end{table}
% 149
% Axie: Plant
% Cards: October Treat | Vegetal Bite | Disguise | Gas Unleash
% Stats: 59 | 31 | 31 | 43
% Axie: Dusk 
% Cards: Barb Strike | Sneaky Raid | Surprise Invasion | Venom Spray
% Stats: 57 | 43 | 27 | 37 
% Axie: Dusk
% Cards: Barb Strike | Chomp | Smart Shot | Venom Spray
% Stats: 51 | 50 | 27 | 36

\begin{table}[H]
	\centering

	\begin{adjustbox}{max width=0.6\textwidth}
	\begin{tabular}{cccc}\hline\hline \noalign{\smallskip}
		\textbf{Team 5} & Axie 1 & Axie 2 & Axie 3  \\
		\noalign{\smallskip}\hline\noalign{\smallskip}
		Axie Class & \text Plant & Dusk & Dusk  \\ %($\pm$ 0.2090)} \\
		\noalign{\smallskip}\hline\noalign{\smallskip}
		Axie Stats & \text 59/31/31/43 & 54/48/27/35 & 51/46/27/40 \\ %($\pm$ 0.2361)\\
		\noalign{\smallskip}\hline\noalign{\smallskip}
		Axie Card 1 & October Treat & Spike Throw & Sticky Goo \\ %($\pm$ 0.2148)   \\
		\noalign{\smallskip}\hline\noalign{\smallskip}
		Axie Card 2 & Vegetal Bite & Chomp & Chomp \\ %($\pm$ 0.2148)   \\
		\noalign{\smallskip}\hline\noalign{\smallskip}
		Axie Card 3 & Disguise & Disarm & Mystic Rush \\ %($\pm$ 0.2148)   \\
		\noalign{\smallskip}\hline\noalign{\smallskip}
		Axie Card 4 & Gas Unleash & Allergic Reaction & Allergic Reaction \\ %($\pm$ 0.2148)   \\

		\noalign{\smallskip}\hline\hline
	\end{tabular}
	\end{adjustbox}

\end{table}
% 169
% Axie: Plant
% Cards: October Treat | Vegetal Bite | Disguise | Gas Unleash
% Stats: 59 | 31 | 31 | 43
% Axie: Dusk
% Cards: Spike Throw | Chomp | Disarm | Allergic Reaction
% Stats: 54 | 48 | 27 | 35
% Axie: Dusk
% Cards: Sticky Goo | Chomp | Mystic Rush | Allergic Reaction
% Stats: 51 | 46 | 27 | 40

\begin{table}[H]
	\centering

	\begin{adjustbox}{max width=0.6\textwidth}
	\begin{tabular}{cccc}\hline\hline \noalign{\smallskip}
		\textbf{Team 6} & Axie 1 & Axie 2 & Axie 3  \\
		\noalign{\smallskip}\hline\noalign{\smallskip}
		Axie Class & \text Plant & Beast & Bird  \\ %($\pm$ 0.2090)} \\
		\noalign{\smallskip}\hline\noalign{\smallskip}
		Axie Stats & \text 59/31/31/43 & 32/45/31/56 & 27/61/35/41 \\ %($\pm$ 0.2361)\\
		\noalign{\smallskip}\hline\noalign{\smallskip}
		Axie Card 1 & October Treat & Single Combat & Blackmail \\ %($\pm$ 0.2148)   \\
		\noalign{\smallskip}\hline\noalign{\smallskip}
		Axie Card 2 & Vegetal Bite & Nut Crack & Dark Swoop \\ %($\pm$ 0.2148)   \\
		\noalign{\smallskip}\hline\noalign{\smallskip}
		Axie Card 3 & Disguise & Nut Throw & Eggbomb \\ %($\pm$ 0.2148)   \\
		\noalign{\smallskip}\hline\noalign{\smallskip}
		Axie Card 4 & Carrot Hammer & Ivory Stab & All-out Shot \\ %($\pm$ 0.2148)   \\

		\noalign{\smallskip}\hline\hline
	\end{tabular}
	\end{adjustbox}

\end{table}

\end{document}